\documentclass[letterpaper, 10 pt, conference]{ieeeconf}  % Comment this line out if you need a4paper

\IEEEoverridecommandlockouts                              % This command is only needed if 
\usepackage{subcaption}
\usepackage{graphicx}

\title{\LARGE \bf
Beyond Pixel Similarity: Task-Aware Evaluation of GAN-Based Synthetic Sonar Data for Robotic Perception
}

\author{Hannan Ejaz Keen$^{1}$, Muhammad Moazam Fraz $^{2}$ and Karsten Berns$^{3}$ % <-this % stops a space
\thanks{$^{1}$Hannan Ejaz Keen is Senior Researcher at XITASO GmbH IT \& Software Solutions, Augsburg, Germany
        {\tt\small hannan.keen@xitaso.com}}%
\thanks{$^{2}$Muhammad Moazam Fraz is Professor in School of Electrical Engineering and Computer Science at National University of Sciences and Technology (NUST), Islamabad, Pakistan
        {\tt\small moazam.fraz@seecs.edu.pk}}%
\thanks{$^{3}$Karsten Berns is Professor in Department of Computer Science at Technical University of Kaiserslautern, Kaiserslautern, Germany
        {\tt\small karsten.berns@cs.rptu.de}}%
}

\begin{document}

\maketitle
\thispagestyle{empty}
\pagestyle{empty}

%%%%%%%%%%%%%%%%%%%%%%%%%%%%%%%%%%%%%%%%%%%%%%%%%%%%%%%%%%%%%%%%%%%%%%%%%%%%%%%%
\begin{abstract}

Synthetic data can reduce the cost of collecting and annotating training data for robotic perception, but generating sensor observations that preserve the characteristics relevant to downstream perception remains challenging, particularly for sonar imagery. In this work, we investigate whether conventional image-fidelity metrics adequately reflect the downstream perception performance of GAN-generated synthetic sonar data. We employ a Pix2Pix conditional generative adversarial network with four discriminator configurations characterized by different receptive fields: PixelGAN, PatchGAN-16, PatchGAN-70, and ImageGAN. The models are trained using sonar imagery from two datasets and evaluated using conventional image-fidelity metrics, including Structural Similarity Index (SSIM), Peak Signal-to-Noise Ratio (PSNR), and Mean Squared Error (MSE). To complement these pixel-level measures with task-oriented evaluation, YOLOX-S, YOLOX-L, and Faster R-CNN detectors are trained exclusively on real sonar imagery and subsequently evaluated on the GAN-generated images using identical test samples and annotations across all discriminator configurations. The results reveal a discrepancy between image-fidelity and downstream object-detection performance: the configuration achieving the best SSIM, PSNR, and MSE does not consistently yield the best detection performance. In particular, PatchGAN configurations achieve strong downstream detection results despite not achieving the highest pixel-level similarity scores. These findings suggest, for the datasets and models considered, pixel-level image-fidelity metrics alone may not consistently capture the task-relevant realism of synthetic sonar observations and motivate the use of task-aware evaluation for synthetic sensor data intended for robotic perception.

\end{abstract}

%%%%%%%%%%%%%%%%%%%%%%%%%%%%%%%%%%%%%%%%%%%%%%%%%%%%%%%%%%%%%%%%%%%%%%%%%%%%%%%%

\section{INTRODUCTION}

Reliable perception in underwater robotics increasingly relies on imaging sonar because acoustic sensing remains effective in environments where optical cameras are degraded by limited illumination, turbidity, and suspended particles. However, sonar imagery exhibits sensor- and environment-dependent characteristics, including speckle noise, target highlights, acoustic shadows, reverberation, and strong variation with range, viewing angle, and seabed conditions. These characteristics make the development of robust sonar perception models challenging and increase the demand for large and diverse annotated datasets. In practice, collecting and annotating such datasets requires specialized equipment and repeated surveys across different environments, making real-world data acquisition costly and time-consuming \cite{lee2018}, \cite{Keen2024}.

\begin{figure}
	\begin{subfigure}{0.49\columnwidth}
		\includegraphics[width=\linewidth]{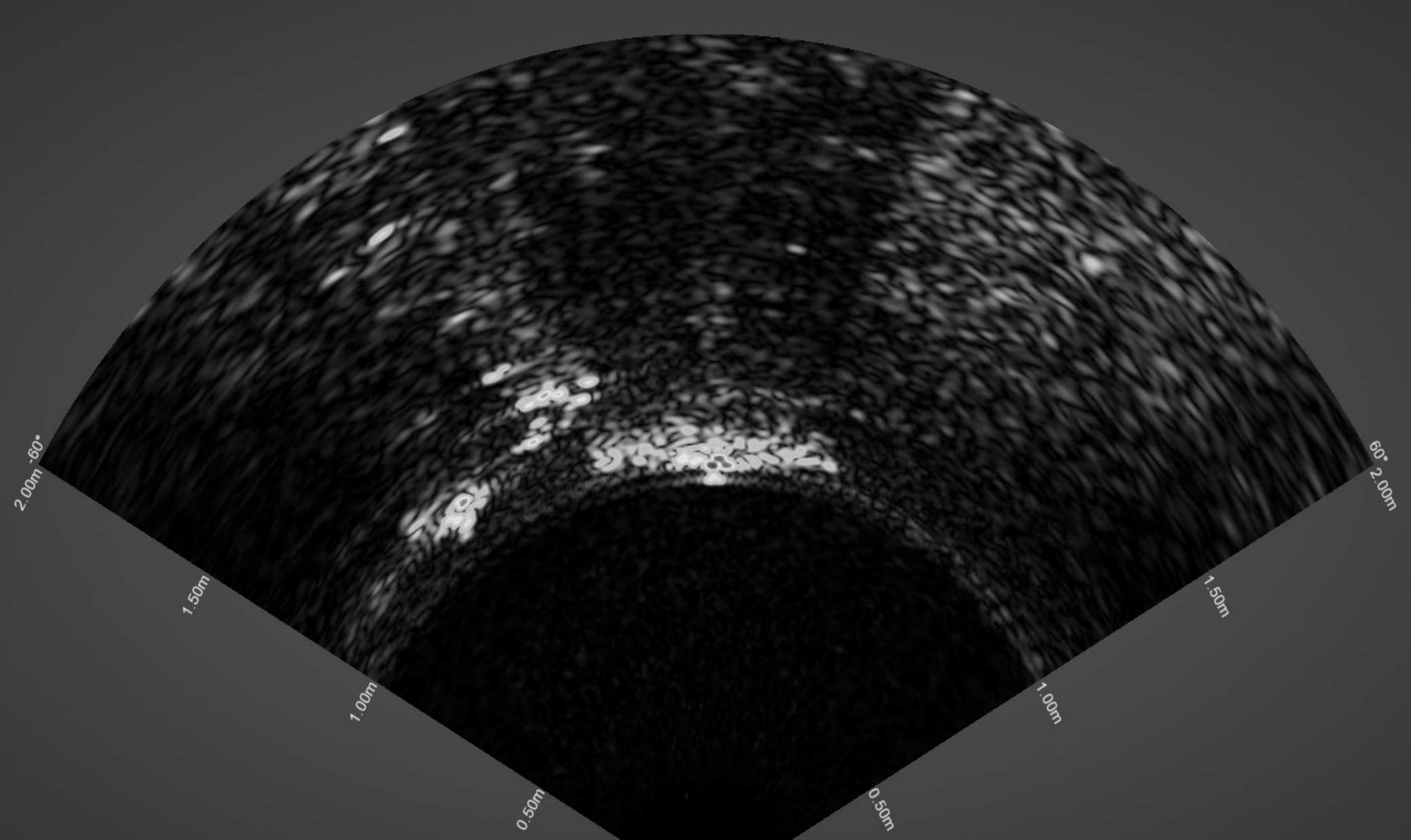}\label{fig:unreal_choi_sonar_a}\caption{}
	\end{subfigure}
	\begin{subfigure}{0.49\columnwidth}
		\includegraphics[width=\linewidth]{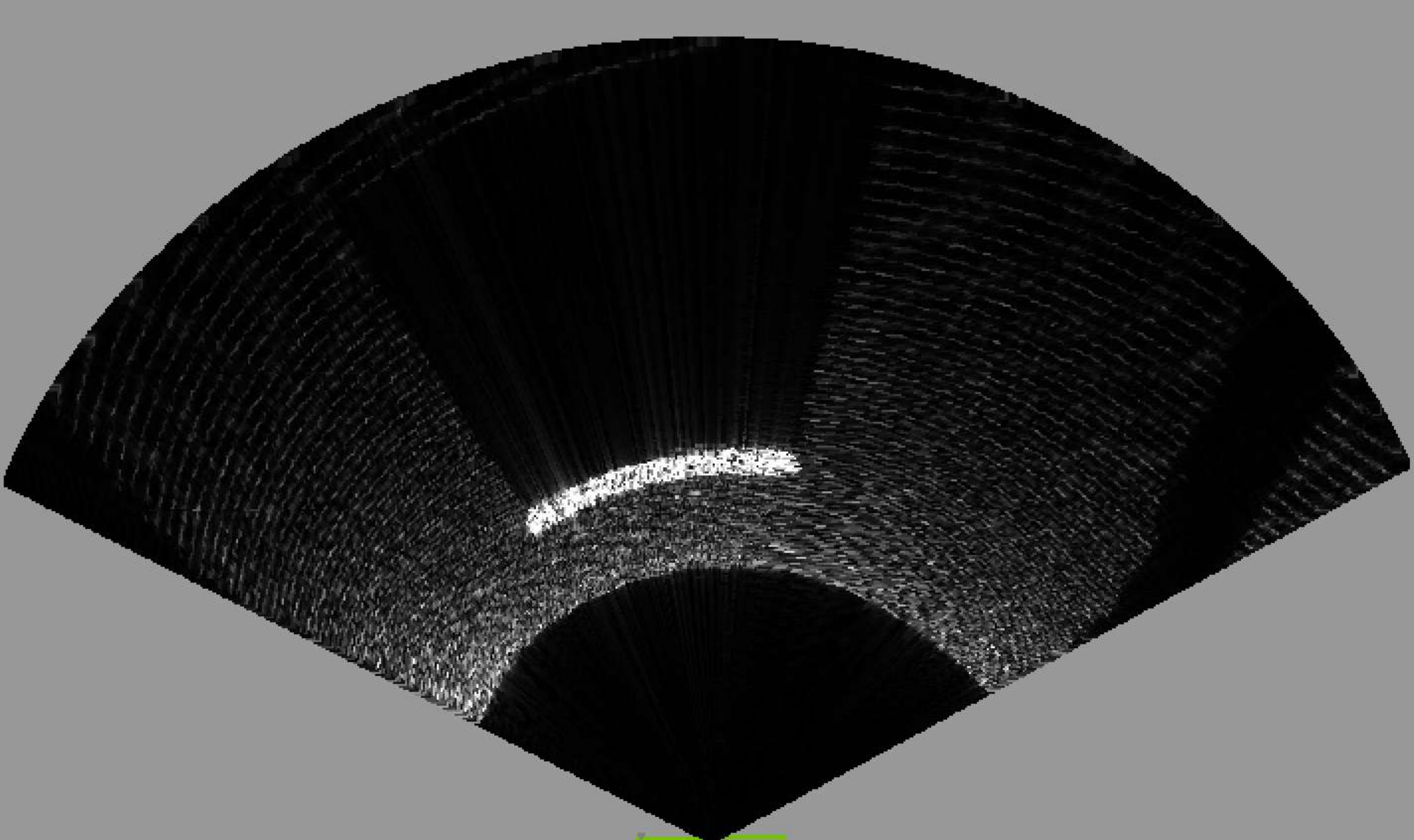}\label{fig:unreal_choi_sonar_b}\caption{}
	\end{subfigure}
	\caption[Comparative visualization between real and simulated sonar imagery.]{Comparative Visualization of Sonar Data: (a) Real sonar image (b) Simulated sonar image \cite{choi_2021}. \label{fig:unreal_choi_sonar}}
\end{figure}

\begin{figure*}[ht!]
        \centering
        %\begin{subfigure}[b]{0.16\textwidth}
            \includegraphics[width=0.95\textwidth]{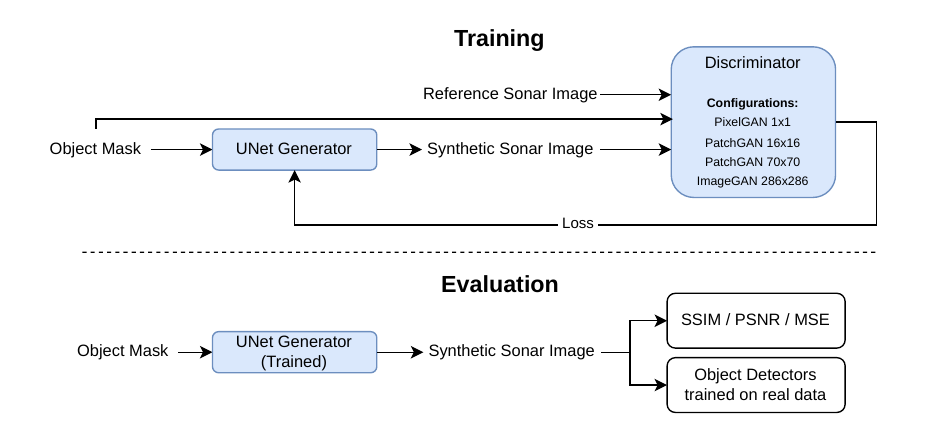}
            \caption{Pix2Pix-based synthetic sonar generation and evaluation pipeline with different discriminator receptive fields}
            \label{fig:pix2pix-pipeline}
    \end{figure*}

Synthetic data provides a potential alternative by enabling large numbers of labeled observations to be generated without repeated real-world acquisition. Previous studies have demonstrated that synthetic sonar observations generated from simulated underwater scenes can support downstream perception \cite{lee2018}. However, conventional sonar simulation can be difficult to model accurately because realistic observations depend on acoustic propagation, scattering, reverberation, sensor characteristics, and environmental interactions as shown in Fig. \ref{fig:unreal_choi_sonar}. Consequently, simulation may compute scene geometry while failing to replicate the appearance characteristics of real sonar imagery, limiting its usefulness for perception.

Generative adversarial networks (GANs) provide one approach for learning this appearance transformation. GAN-based methods have been applied to generate or translate sonar imagery from simulated or conditioned representations and have demonstrated potential for downstream underwater perception \cite{Sung2020-gt}, \cite{noren2021}, \cite{Jegorova2020}. These studies motivate the use of learned image translation for synthetic sonar generation. However, they also raise a fundamental research question:

\begin{itemize}
    \item [RQ:] \textbf{Does image similarity to real sonar data indicate how useful synthetic sonar data is for robotic perception?}
\end{itemize}

Synthetic images are commonly evaluated using metrics such as Structural Similarity Index (SSIM), Peak Signal-to-Noise Ratio (PSNR), and Mean Squared Error (MSE). Although these metrics quantify similarity to a reference image, they primarily measure pixel-level or local structural features \cite{keen_2023}. They do not directly assess whether task-relevant characteristics, such as target boundaries, acoustic shadows, and local contrast patterns, are preserved sufficiently for object detection. Prior work on synthetic data and sonar perception has similarly highlighted the importance of evaluating generated data according to downstream task performance rather than visual similarity alone \cite{Liu_2019_ICCV}-\cite{Ma2020-wacv}. %This suggests that image fidelity and task utility should be considered complementary rather than interchangeable evaluation criteria.

This issue is particularly relevant in conditional image-to-image translation. In Pix2Pix \cite{pix2pix2017}, the discriminator evaluates image realism at a particular spatial scale determined by its receptive field. A pixel-level discriminator focuses on local intensity statistics, while PatchGAN discriminators evaluate increasingly larger local regions; an image-level discriminator instead evaluates broader spatial consistency. These different receptive fields may therefore influence whether the generator prioritizes local structures or global appearance characteristics.

In this work, we investigate whether conventional image-fidelity metrics adequately reflect the downstream perception performance of GAN-generated synthetic sonar data. Within a Pix2Pix framework, we compare four discriminator configurations: PixelGAN, PatchGAN-16, PatchGAN-70, and ImageGAN. The generated images are evaluated using SSIM, PSNR, and MSE and, independently through YOLOX-S, YOLOX-L, and Faster R-CNN detectors. The detectors are trained exclusively on real sonar imagery and subsequently evaluated on the generated images, with identical test samples and annotations used across all discriminator configurations. The experiments were originally conducted as part of the authors' doctoral research \cite{Keen2024} and are revisited here from the perspective of task-aware synthetic-sensor evaluation.

Our results reveal a discrepancy between image-fidelity evaluation and downstream detection performance in the experimental settings considered. The discriminator configuration achieving the strongest pixel-level fidelity does not consistently provide the strongest downstream detection performance. In particular, PatchGAN configurations frequently achieve higher detector performance despite lower SSIM, PSNR, or MSE values. These findings indicate that pixel-level image-fidelity metrics alone may not consistently characterize the task-relevant realism of synthetic sonar observations.

\section{Methodology}
\subsection{Sonar Datasets and Conditioning}
We use two sonar datasets with annotations suitable for conditional image generation and downstream object detection: the Marine Debris dataset \cite{singh_2021} and the Underwater Acoustic Target Detection (UATD) dataset \cite{xie_2022}. The datasets differ in acquisition conditions and therefore provide distinct sonar-image characteristics. The Marine Debris data is acquired in a controlled pool environment, while UATD contains imagery acquired in a more variable river environment.

For conditional image generation, object masks are used as the input representation. The Marine Debris dataset provides segmentation masks. UATD provides bounding-box annotations; therefore, binary masks are generated by assigning pixels inside each annotated bounding box to the object class and the remaining pixels to the background. This provides a common conditioning representation for both datasets.

Because the source sonar data is acquired as video sequences, consecutive frames can contain substantial redundancy. Similar images are therefore filtered during preprocessing. %using SSIM, PSNR, and MSE thresholds
In addition, data augmentation is investigated to improve generation robustness; mirroring and random jitter are retained, while other tested transformations introduce undesirable artifacts \cite{Keen2024}.

\subsection{Pix2Pix-Based Sonar Image Generation}
We employ a conditional GAN based on Pix2Pix \cite{pix2pix2017}. The generator uses a U-Net architecture, while the discriminator is varied to investigate the effect of receptive field. Across all experiments, the generator architecture, training data, preprocessing, and optimization protocol are kept fixed. An overview of the experimental pipeline is shown in Fig. \ref{fig:pix2pix-pipeline}.

The conditional adversarial objective is combined with a ($L_1$) reconstruction loss:

\begin{equation}
    G^* = \arg\min_G \max_D L_{cGAN}(G,D) + \lambda L_{L1}(G)
\end{equation}

where the adversarial term encourages the generated sonar image to appear consistent with the target domain and the ($L1$) term encourages correspondence with the paired reference image \cite{pix2pix2017}.

Four discriminator configurations are evaluated (Fig. \ref{fig:pix2pix-pipeline}):
\begin{itemize}
\item \textbf{PixelGAN:} pixel-level discrimination with a minimal receptive field;
\item \textbf{PatchGAN-16:} discrimination over local ($16\times16$) regions;
\item \textbf{PatchGAN-70:} discrimination over local ($70\times70$) regions;
\item \textbf{ImageGAN:} discrimination over the full image, corresponding to a ($286\times286$) receptive field.
\end{itemize}
This controlled design isolates the effect of discriminator spatial scale on the generated sonar imagery.

\subsection{Evaluation Protocol}
The generated images are evaluated from two complementary perspectives.

\textbf{Image fidelity:} SSIM, PSNR, and MSE are computed between each generated image and its corresponding reference sonar image. Higher SSIM and PSNR and lower MSE indicate greater image-level similarity.

\textbf{Task-level evaluation:} YOLOX-S, YOLOX-L, and Faster R-CNN are trained exclusively on real sonar imagery. The trained detectors are then applied to generated sonar images from each discriminator configuration. The same test samples and annotations are used for all four configurations. We report Average Precision (AP) averaged over IoU thresholds from 0.50 to 0.95, together with recall. We denote this metric as AP throughout the paper. The detectors are trained only on real sonar data and then tested on the generated images. This allows us to assess how useful the synthetic images are for detectors trained on real sonar data.

We refer to this property as \textbf{task-relevant realism}, meaning the extent to which a generated sonar image preserves features useful for downstream detection. This term does not imply complete physical or sensor-level realism.

\begin{table}[h]
    	\centering
    	\caption[]{Image-fidelity performance of discriminator configurations. Best result in each metric/dataset is highlighted.}
    	\label{tab:gans_direct_comparison}
    	\resizebox{\columnwidth}{!}{%
        \begin{tabular}{ | c || c c c | c c c | }
            \hline
    		\textbf{Discriminator} & \multicolumn{3}{c|}{\textbf{Marine Debris Dataset}} & \multicolumn{3}{c|}{\textbf{UATD Dataset}} \\ 
    		\hline
    		& \textit{SSIM} & \textit{PSNR} & \textit{MSE} & \textit{SSIM} & \textit{PSNR} & \textit{MSE} \\
    		PixelGAN & \textbf{0.586} & \textbf{24.265} & \textbf{49.532} & 0.755 & 31.845 & 21.078 \\
    		PatchGAN-16 & 0.546 & 23.090 & 52.079 & 0.729 & 30.995 & 23.759\\
    		PatchGAN-70 & 0.561 & 23.558 & 51.209 & 0.749 & 31.331 & 23.014\\
    		ImageGAN & 0.548 & 23.192 & 52.174 & \textbf{0.760} & \textbf{32.034} & \textbf{20.800}\\
    		\hline
    	\end{tabular}
        }
\end{table}

\section{Results and Discussion}
\subsection{Image-Fidelity Results}
Table \ref{tab:gans_direct_comparison} summarizes the image-fidelity performance on the Marine Debris and UATD datasets.

The results show that the preferred discriminator depends on the dataset. PixelGAN obtains the strongest SSIM, PSNR, and MSE results on Marine Debris, whereas ImageGAN achieves the strongest values on UATD. Thus, conventional image-fidelity metrics would favor different discriminator configurations depending on the dataset.

\begin{figure*}
        \centering
        \begin{subfigure}[b]{\textwidth}
        \begin{subfigure}[b]{0.16\textwidth}
            \includegraphics[width=\textwidth]{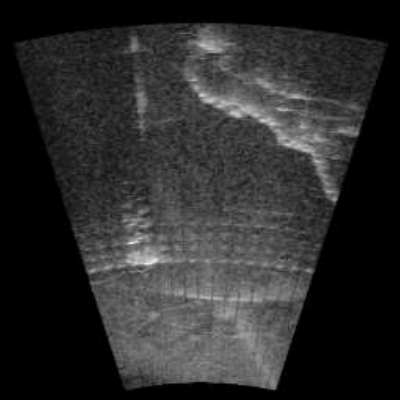}
            \caption*{Reference}
            \label{fig:marine_debris_patch_size_comparison_gt}
        \end{subfigure}
        \begin{subfigure}[b]{0.16\textwidth}
            \includegraphics[width=\textwidth]{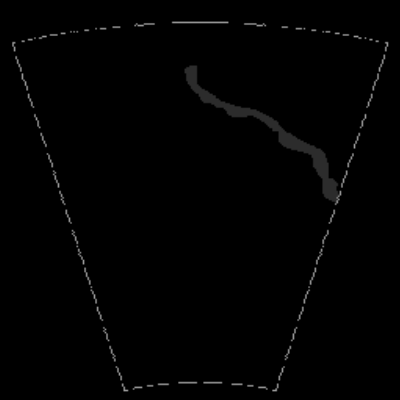}
            \caption*{Mask}
            \label{fig:marine_debris_patch_size_comparison_mask}
        \end{subfigure}
        \begin{subfigure}[b]{0.16\textwidth}
            \includegraphics[width=\textwidth]{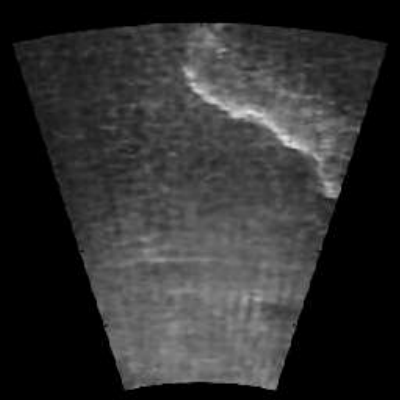}
            \caption*{PixelGAN}
            \label{fig:marine_debris_patch_size_comparison_ps1}
        \end{subfigure}
        \begin{subfigure}[b]{0.16\textwidth}
            \includegraphics[width=\textwidth]{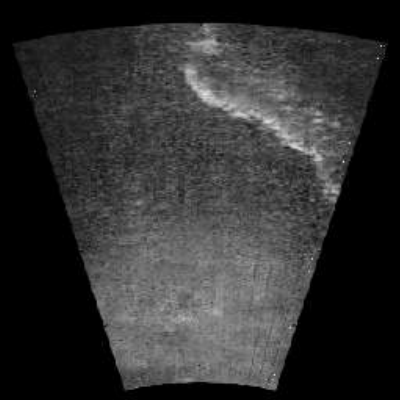}
            \caption*{PatchGAN-16}
            \label{fig:marine_debris_patch_size_comparison_ps16}
        \end{subfigure}
        \begin{subfigure}[b]{0.16\textwidth}
            \includegraphics[width=\textwidth]{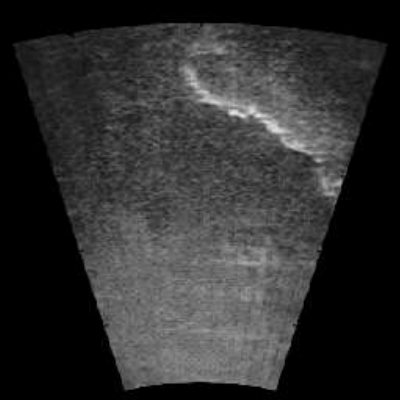}
            \caption*{PatchGAN-70}
            \label{fig:marine_debris_patch_size_comparison_ps70}
        \end{subfigure}
        \begin{subfigure}[b]{0.16\textwidth}
            \includegraphics[width=\textwidth]{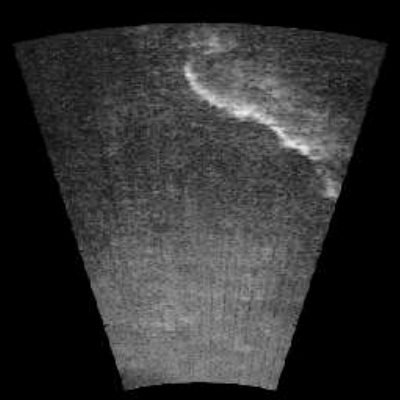}
            \caption*{ImageGAN}
            \label{fig:marine_debris_patch_size_comparison_ps256}
        \end{subfigure}
        \caption[Marine Debris Dataset]{Marine Debris Dataset}
        \label{fig:marine_debris_patch_size_comparison}
        \end{subfigure}

        \begin{subfigure}[b]{\textwidth}
        \begin{subfigure}[b]{0.16\textwidth}
            \includegraphics[width=\textwidth]{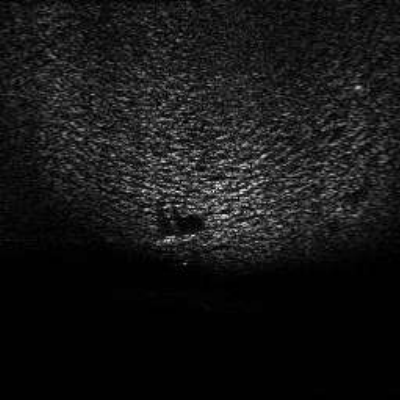}
            \caption*{Reference}
            \label{fig:uatd_patch_size_comparison_gt}
        \end{subfigure}
        \begin{subfigure}[b]{0.16\textwidth}
            \includegraphics[width=\textwidth]{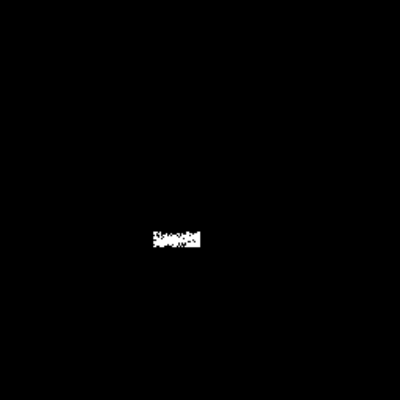}
            \caption*{Mask}
            \label{fig:uatd_patch_size_comparison_mask}
        \end{subfigure}
        \begin{subfigure}[b]{0.16\textwidth}
            \includegraphics[width=\textwidth]{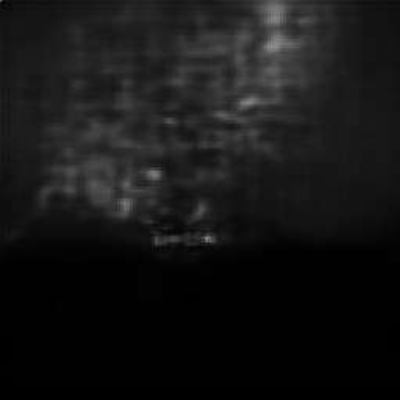}
            \caption*{PixelGAN}
            \label{fig:uatd_patch_size_comparison_ps1}
        \end{subfigure}
        \begin{subfigure}[b]{0.16\textwidth}
            \includegraphics[width=\textwidth]{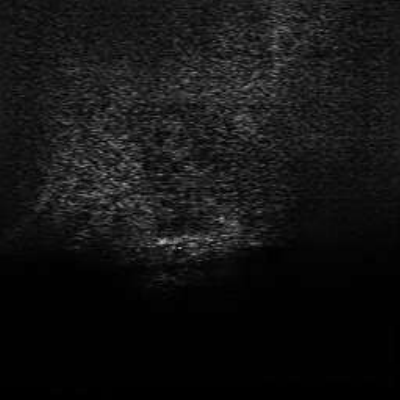}
            \caption*{PatchGAN-16}
            \label{fig:uatd_patch_size_comparison_ps16}
        \end{subfigure}
        \begin{subfigure}[b]{0.16\textwidth}
            \includegraphics[width=\textwidth]{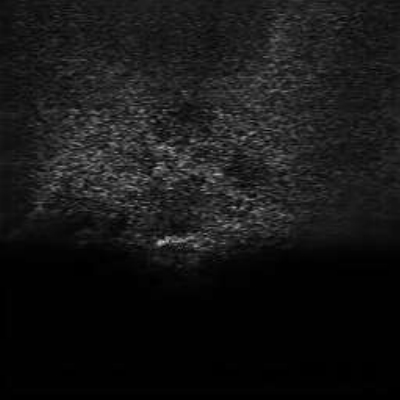}
            \caption*{PatchGAN-70}
            \label{fig:uatd_patch_size_comparison_ps70}
        \end{subfigure}
        \begin{subfigure}[b]{0.16\textwidth}
            \includegraphics[width=\textwidth]{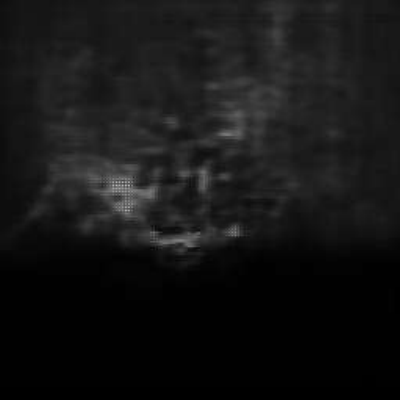}
            \caption*{ImageGAN}
            \label{fig:uatd_patch_size_comparison_ps256}
        \end{subfigure}
        \caption[UATD Dataset]{UATD Dataset}
        \label{fig:uatd_patch_size_comparison}
        \end{subfigure}
        \caption[Qualitative comparison of generated sonar images across discriminator configurations and datasets.]{Qualitative comparison of generated sonar images across discriminator configurations and datasets.}
        \label{fig:qualitative-results}
    \end{figure*}

However, numerical image similarity does not fully describe the observed appearance. Fig. \ref{fig:qualitative-results} shows that configurations with strong pixel-level scores can generate smoother or blurrier images, whereas PatchGAN configurations can preserve sharper local structures. 

For the Marine Debris dataset, PixelGAN achieves the highest SSIM and PSNR among the evaluated configurations. However, as shown in Fig. \ref{fig:marine_debris_patch_size_comparison}, its generated images appear comparatively blurred, whereas PatchGAN produces sharper local structures. Similarly, for the UATD dataset, although ImageGAN achieves the highest SSIM, PSNR, and lowest MSE, Fig. \ref{fig:uatd_patch_size_comparison} shows that the ImageGAN-generated image contains fewer details than those produced by the PatchGAN configurations.

These observations suggest that conventional pixel-level fidelity metrics do not necessarily reflect the characteristics of synthetic sonar images that are most relevant for downstream perception.

\subsection{Downstream Detection Results}
Table \ref{tab:gans_indirect_comparison} reports detector performance on generated sonar images. The detectors were trained only on real sonar imagery and then evaluated on the generated images.

\begin{table}[h]
	\centering
	\caption[]{Downstream object-detection performance on generated sonar imagery. Entries show AP@[0.50:0.95] / Recall.}
	\label{tab:gans_indirect_comparison}
    \resizebox{\columnwidth}{!}{%
	\begin{tabular}{ | c || c | c | c | }
        \hline
		\textbf{Discriminator} &{\textbf{YOLOX-S}} & {\textbf{YOLOX-L}} & {\textbf{Faster R-CNN}}\\ 
        \hline
        \multicolumn{4}{c}{\textbf{Marine Debris Dataset}}\\
        \hline
		PixelGAN & 0.76 / 0.54 & 0.67 / 0.46 & 0.66 / 0.43 \\
		\textbf{PatchGAN-16} & \textbf{0.81} / \textbf{0.56} & \textbf{0.76} / \textbf{0.51} & 0.66 / 0.43 \\
		PatchGAN-70 & 0.81 / 0.55 & 0.72 / 0.48 & \textbf{0.67} / \textbf{0.42}\\
		ImageGAN & 0.79 / 0.56 & 0.71 / 0.49 & 0.61 / 0.45 \\
		\hline
        \multicolumn{4}{c}{\textbf{UATD Dataset}} \\
        \hline
        PixelGAN & 0.62 / 0.29 & 0.61 / 0.30 & 0.55 / 0.25 \\
		PatchGAN-16 & 0.67 / 0.33 & 0.71 / 0.36 & 0.54 / 0.26\\
		\textbf{PatchGAN-70} & \textbf{0.72} / \textbf{0.34} & \textbf{0.74} / \textbf{0.36} & 0.58 / 0.29 \\
		ImageGAN & 0.68 / 0.34 & 0.69 / 0.33 & \textbf{0.64} / \textbf{0.32}\\
		\hline
	\end{tabular}
    }
\end{table}

The downstream results reveal a different ranking from the image-fidelity metrics. On Marine Debris, PixelGAN achieves the highest SSIM, PSNR, and lowest MSE, but PatchGAN-16 and PatchGAN-70 obtain higher YOLOX-S AP, while PatchGAN-16 obtains the highest YOLOX-L AP. On UATD, ImageGAN provides the strongest image-fidelity scores, yet PatchGAN-70 obtains the highest YOLOX-S and YOLOX-L AP.

The discrepancy is particularly clear when comparing the best image-level and task-level configurations. For UATD, ImageGAN achieves SSIM = 0.760, PSNR = 32.034 dB, and MSE = 20.800, but PatchGAN-70 achieves higher YOLOX-S and YOLOX-L AP values of 0.72 and 0.74, respectively. Similarly, on Marine Debris, PixelGAN has the strongest image-fidelity scores, whereas PatchGAN-16 achieves the strongest YOLOX-S and YOLOX-L AP.

These results indicate, in our experiments, maximizing similarity to a reference image does not necessarily maximize the preservation of features required by a detector trained on real sonar imagery.

\subsection{Discussion: Image Fidelity Versus Task-Relevant Realism}
The observed mismatch can be explained by the different properties measured by the two evaluation approaches. SSIM, PSNR, and MSE reward close correspondence to the reference image at the pixel or local-structure level. A generated image may therefore obtain a favorable score by reproducing overall intensity and structural statistics while smoothing or modifying small target structures.

In contrast, the detector-based evaluation is sensitive to whether generated images retain cues that are discriminative for object localization. The stronger downstream performance of PatchGAN configurations suggests that image characteristics beyond pointwise similarity are important for synthetic sonar perception. One possible explanation is that discriminator configurations operating over local regions encourage image characteristics that are more useful to the detector, although this mechanism is not directly isolated in our experiments. However, the results do not indicate a universally optimal discriminator: the preferred configuration varies with dataset and detector.

The main finding is therefore not that one particular discriminator is universally superior, but that the ranking of generation methods depends on the evaluation criterion. These results suggest that image-fidelity metrics should be complemented with task-aware evaluation when synthetic sonar imagery is intended for robotic perception.

Importantly, the detector transfer experiment provides evidence of task-relevant realism, not a complete measurement of physical sensor realism. Likewise, the present study does not quantify a reduction of the simulator-to-real domain gap because it does not directly compare raw simulator outputs with real sonar under the same transfer protocol. Instead, it demonstrates that different metrics of synthetic-image quality can lead to different conclusions about the same generated data.

\section{Conclusion}
This work investigates whether conventional image-fidelity metrics adequately characterize the task-relevant realism of GAN-generated synthetic sonar imagery. Within a Pix2Pix framework, four discriminator configurations (PixelGAN, PatchGAN-16, PatchGAN-70, and ImageGAN) are evaluated using SSIM, PSNR, and MSE as well as downstream object detection with YOLOX-S, YOLOX-L, and Faster R-CNN are trained exclusively on real sonar data.

Across the two sonar datasets considered, the configuration achieving the strongest image-fidelity scores does not consistently achieve the strongest object-detection performance. PatchGAN configurations frequently produce stronger downstream detection despite lower pixel-level similarity. These results indicate that SSIM, PSNR, and MSE capture only part of the quality relevant to synthetic sonar perception and should therefore be complemented with task-aware evaluation.

For synthetic sonar data intended for robotic perception, these results support evaluating data according to its intended use rather than image similarity alone. Future work should extend the analysis to additional sonar sensors and datasets and directly compare raw simulator outputs, generated observations, and real sonar to quantify broader domain-gap effects.

\bibliographystyle{IEEEtran}
\bibliography{ref.bib}

\end{document}